\documentclass{article}

\usepackage{PRIMEarxiv}
\usepackage{amsmath}
\usepackage{multirow}
\usepackage[utf8]{inputenc} 
\usepackage[T1]{fontenc}    
\usepackage{hyperref}       
\usepackage{url}            
\usepackage{booktabs}       
\usepackage{amsfonts}       
\usepackage{nicefrac}       
\usepackage{microtype}      
\usepackage{lipsum}
\usepackage{fancyhdr}       
\usepackage{graphicx}       
\graphicspath{{media/}}     

\title{M-Gaussian: An Magnetic Gaussian Framework for Efficient Multi-Stack MRI Reconstruction
\thanks{(Corresponding authors: Mengting Liu and Liangqiong Qu.)}
\thanks{K. Zheng, X. Cai, J. Wang, G. Fu, Z. Li and M. Liu are with School of Biomedical Engineering, Shenzhen Campus of Sun Yat-sen University, Shenzhen, 518107, China. (e-mail: \{zhengky29, caix63, wangjq236, fugx3\}@mail2.sysu.edu.cn; \{lizhsh36, liumt55\}@mail.sysu.edu.cn).}
\thanks{Y. Chen and X. Ge are with the School of Information Science and Engineering, Shandong Normal University, Shandong, China. (e-mail: yachauchen@gmail.com; xintingge@sdnu.edu.cn).}
\thanks{L. Qu is with the School of Computing and Data Science, The University of Hong Kong, Hong Kong, China. (e-mail: liangqqu@hku.hk).}
}

\author{
 \textnormal{Kangyuan Zheng, Xuan Cai, Jiangqi Wang, Guixing Fu, Zhuoshuo Li,} \\ Yazhou Chen, Xinting Ge, Liangqiong Qu, Mengting Liu
}

\begin{document}
\maketitle

\begin{abstract}
Magnetic Resonance Imaging (MRI) is a crucial non-invasive imaging modality. In routine clinical practice, multi-stack thick-slice acquisitions are widely used to reduce scan time and motion sensitivity, particularly in challenging scenarios such as fetal brain imaging. However, the resulting severe through-plane anisotropy compromises volumetric analysis and downstream quantitative assessment, necessitating robust reconstruction of isotropic high-resolution volumes. Implicit neural representation methods, while achieving high quality, suffer from computational inefficiency due to complex network structures. We present M-Gaussian, adapting 3D Gaussian Splatting to MRI reconstruction. Our contributions include: (1) Magnetic Gaussian primitives with physics-consistent volumetric rendering, (2) neural residual field for high-frequency detail refinement, and (3) multi-resolution progressive training. Our method achieves an optimal balance between quality and speed. On the FeTA dataset, M-Gaussian achieves 40.31 dB PSNR while being 14 times faster, representing the first successful adaptation of 3D Gaussian Splatting to multi-stack MRI reconstruction.
\end{abstract}


\begin{figure*}[t]
\centering
\includegraphics[width=0.8\textwidth]{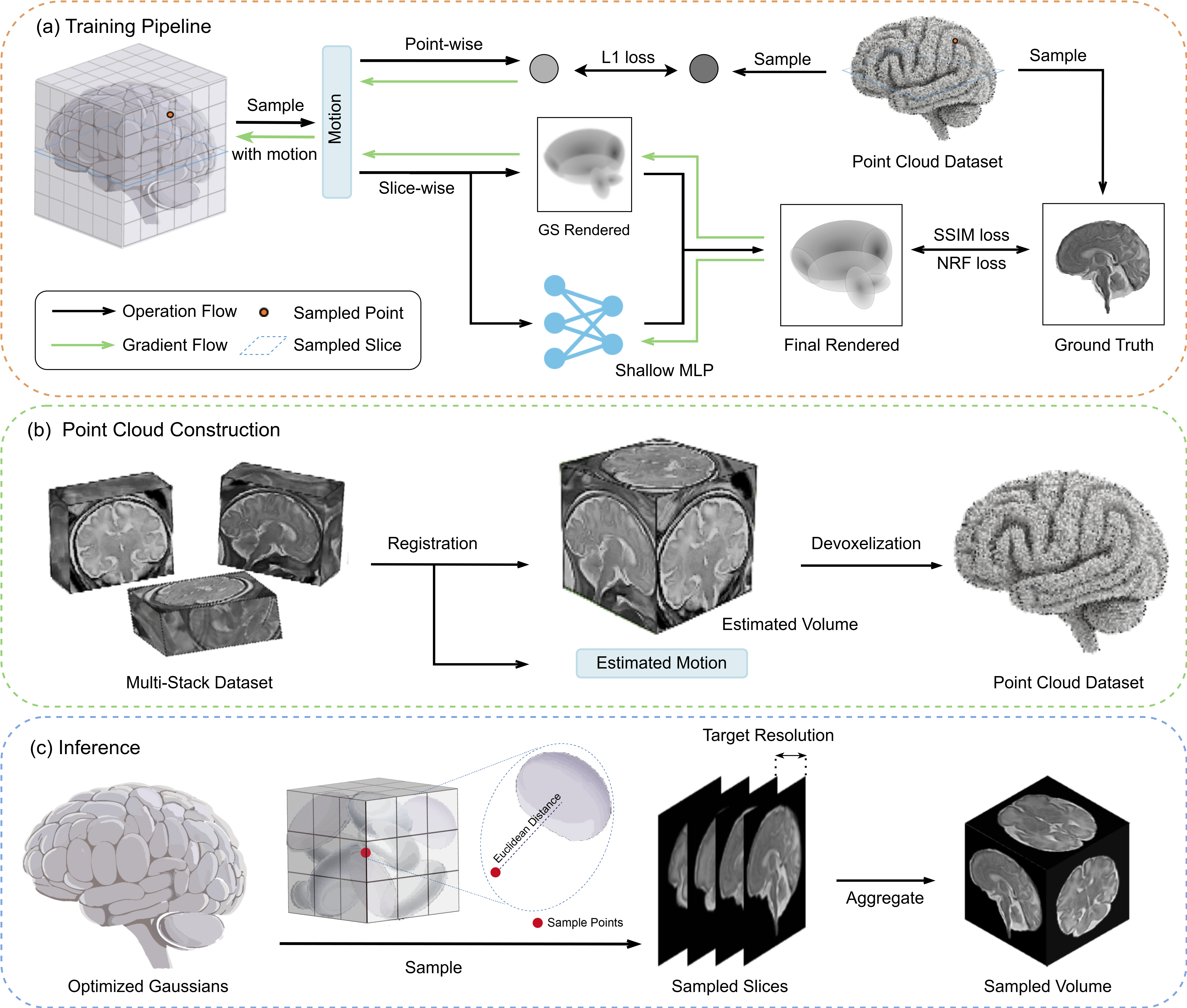}
\caption{Overview of the M-Gaussian framework. 
(a) The training pipeline integrates Gaussian rendering with a Neural Residual Field (NRF) refinement module. 
(b) Point cloud construction is performed via multi-stack registration and devoxelization to initialize the Gaussians. 
(c) The inference stage generates the final volume through Gaussian sampling and aggregation at the target resolution.}

\label{fig_ppl}
\end{figure*}

\section{Introduction}
\label{sec:introduction}
Magnetic Resonance Imaging (MRI) stands as one of the most important non-invasive imaging modalities in modern medicine, providing exceptional soft tissue contrast and multi-parametric information crucial for diagnosis and treatment planning~\cite{mriprinciplesandapplications}. Ideally, MRI would yield high-resolution isotropic 3D volumes with uniform spatial resolution in all dimensions, enabling arbitrary multi-planar reformatting and precise volumetric measurements. However, fundamental physical constraints create an unavoidable trade-off between spatial resolution, signal-to-noise ratio (SNR), and scan time~\cite{timetradeoff}. Direct acquisition of isotropic high-resolution volumes requires prohibitively long scan times, making it impractical for many clinical scenarios, particularly fetal imaging where unpredictable motion demands ultra-fast acquisition protocols, and adult brain imaging where patient comfort and clinical workflow efficiency are essential.

Consequently, clinical practice has adopted a pragmatic compromise: acquiring multiple stacks of thick 2D slices from different orientations~\cite{mrithickslicerecon}. This multi-stack acquisition strategy significantly reduces scan time while maintaining high in-plane resolution, but introduces severe anisotropy with slice thickness often 3-5 times larger than in-plane resolution. The resulting data suffers from partial volume effects~\cite{partialvolumeeffect}, inter-slice gaps, and potential misalignment due to patient motion~\cite{mriartifacts}. These artifacts significantly compromise quantitative analysis capabilities, making the reconstruction of high-quality isotropic volumes from such anisotropic multi-stack data a critical problem in medical imaging.

Traditional slice-to-volume reconstruction methods~\cite{niftymic} have approached this problem through iterative optimization frameworks with explicit regularization terms. While achieving clinical success, these approaches are computationally expensive and scale poorly for high-resolution reconstruction. 

The emergence of implicit neural representation (INR) methods has introduced a paradigm shift~\cite{wang2024neural}. By parameterizing the volume as a continuous function through coordinate-based neural networks, INR methods offer resolution-agnostic reconstruction and the ability to learn complex tissue priors directly from data. However, INR relies on coordinate-based multi-layer perceptrons (MLPs), creating a computational bottleneck where each spatial query necessitates a complete forward pass through the network, resulting in inherently long training and inference times.

In contrast, recent advances in 3D Gaussian Splatting (3DGS)~\cite{3dgs} have demonstrated that explicit primitive-based representations can substantially accelerate volumetric reconstruction without sacrificing quality. By representing scenes as collections of anisotropic 3D Gaussians with learnable parameters, 3DGS enables efficient rendering through direct primitive evaluation and tile-based rasterization rather than substantial network queries. The remarkable success of 3DGS in computer vision—from novel view synthesis to dynamic scene reconstruction—strongly motivates exploring similar benefits for MRI reconstruction from thick-slice acquisitions.

However, adapting 3DGS to MRI presents fundamental challenges. The imaging physics of MRI fundamentally differs from optical imaging: MRI signals represent volumetric tissue properties rather than surface reflectance, requiring complete redesign of the Gaussian properties and rendering pipeline. Furthermore, the volumetric nature of MRI reconstruction requires evaluating millions of 3D points rather than projecting onto 2D image planes, making the projection-based rendering of standard 3DGS computationally prohibitive.

We present MRI-tailored 3D Gaussian (M-Gaussian), the first successful adaptation of 3DGS for MRI reconstruction from multi-stack thick-slice acquisitions. Our method addresses these fundamental challenges through a comprehensive framework combining the efficiency of explicit Gaussian representations with domain-specific innovations tailored for MRI. Our key contributions include:
\begin{itemize}
\item Magnetic Gaussian primitives representing tissue-specific signal intensities with a volumetric rendering pipeline consistent with MRI physics.
\item Block-based spatial partitioning that restricts Gaussian evaluation to local neighborhoods for efficient querying.
\item A neural residual field that complements the smooth Gaussian representation by capturing high-frequency anatomical details.
\item A multi-resolution progressive training strategy that ensures stable convergence for high-resolution reconstruction.
\end{itemize}

\section{Related Work}
\label{sec:related}

\subsection{Multi-Stack MRI Reconstruction} 

Reconstructing isotropic volumes from anisotropic multi-stack MRI data has been extensively studied.
Early approaches~\cite{rousseau2006registration} pioneered registration-based techniques using scattered data interpolation, establishing foundational methodologies. While ~\cite{gholipour2010robust} formulated the problem as robust super-resolution with M-estimation for outlier handling. Building upon these, ~\cite{svrtk} introduced complete outlier removal using robust statistics based on expectation maximization with intensity matching for bias field correction. Subsequently, ~\cite{mrithickslicerecon} proposed multi-level B-spline interpolation, and ~\cite{niftymic} developed an automated framework integrating brain localization, segmentation, and slice-level outlier rejection.

Recent advances have incorporated GPU acceleration~\cite{kainz2015fast} and total variation regularization~\cite{tourbier2015efficient} to enhance computational efficiency and reconstruction quality. Nevertheless, these explicit approaches remain constrained by iterative optimization schemes, exhibiting computational complexity that scales poorly with increasing output resolution.

\subsection{Neural Representations in Medical Imaging} 

Neural Radiance Fields have emerged as a powerful paradigm for 3D reconstruction and rendering. Recent works have successfully adapted Neural Radiance Fields~\cite{mildenhall2021nerf} to various medical imaging modalities: MedNeRF~\cite{corona2022mednerf} for X-ray imaging and Ultra-NeRF~\cite{wysocki2024ultra} for ultrasound imaging.
In MRI imaging, NeSVoR~\cite{xu2023nesvor} demonstrates the potential of INR for slice-to-volume reconstruction in fetal MRI, achieving resolution-agnostic reconstruction. 
IREM~\cite{wu2021irem} similarly applied INR to super-resolution reconstruction of adult brain MRI.

While achieving resolution-agnostic reconstruction, these coordinate-based MLP approaches require numerous network evaluations, resulting in inherently slow training and inference without dedicated implementation optimizations.

\subsection{3D Gaussian Splatting}
3D Gaussian Splatting represents scenes as collections of 3D Gaussian primitives, enabling real-time rendering through efficient rasterization while achieving superior reconstruction quality with significantly faster training compared to neural radiance fields.
The method has been successfully extended to various domains including dynamic scenes~\cite{wu20244d,huang2024sc,guo2024motion} and surface reconstruction~\cite{chen2023neusg, lyu20243dgsr}.

More recently, researchers have begun exploring 3DGS applications in medical imaging. Several works have adapted the framework for X-ray-based imaging: X-Gaussian~\cite{xgaussian} for X-ray novel view synthesis, $\mathrm{R}^2$-Gaussian~\cite{r2gaussian} and 3DGR-CT~\cite{3dgrct} for tomographic reconstruction, and $\mathrm{x}^2$-Gaussian~\cite{x2gaussian} for continuous-time tomographic reconstruction. However, these methods are designed for X-ray imaging with its specific projection model and do not address the unique challenges of MRI thick-slice acquisition and soft tissue contrast.

\section{Method}
\label{sec:method}

\subsection{Overall Pipeline}

Fig.~\ref{fig_ppl} depicts our training and inference pipeline. We construct a unified point cloud by registering input slice stacks into a common RAS (Right-Anterior-Superior) anatomical space and sampling foreground pixels. Each sample consists of a 3D coordinate normalized to the canonical $[-1, 1]^3$ space and its corresponding intensity value. 

We initialize a uniform grid of 3D Gaussians at low resolution and train through differentiable rendering. The model is optimized using the Adam optimizer with point-wise, structural, and regularization losses. To capture fine anatomical details beyond the Gaussian representation's capacity, a lightweight Neural Residual Field is incorporated in subsequent training stages. 

Upon convergence, the learned continuous representation—combining both Gaussian primitives and neural residuals—is sampled at target voxel coordinates to produce the final high-resolution volume.

\subsection{Magnetic 3D Gaussian Representation}

\begin{figure}[t]
\centering
\includegraphics[width=0.7\columnwidth]{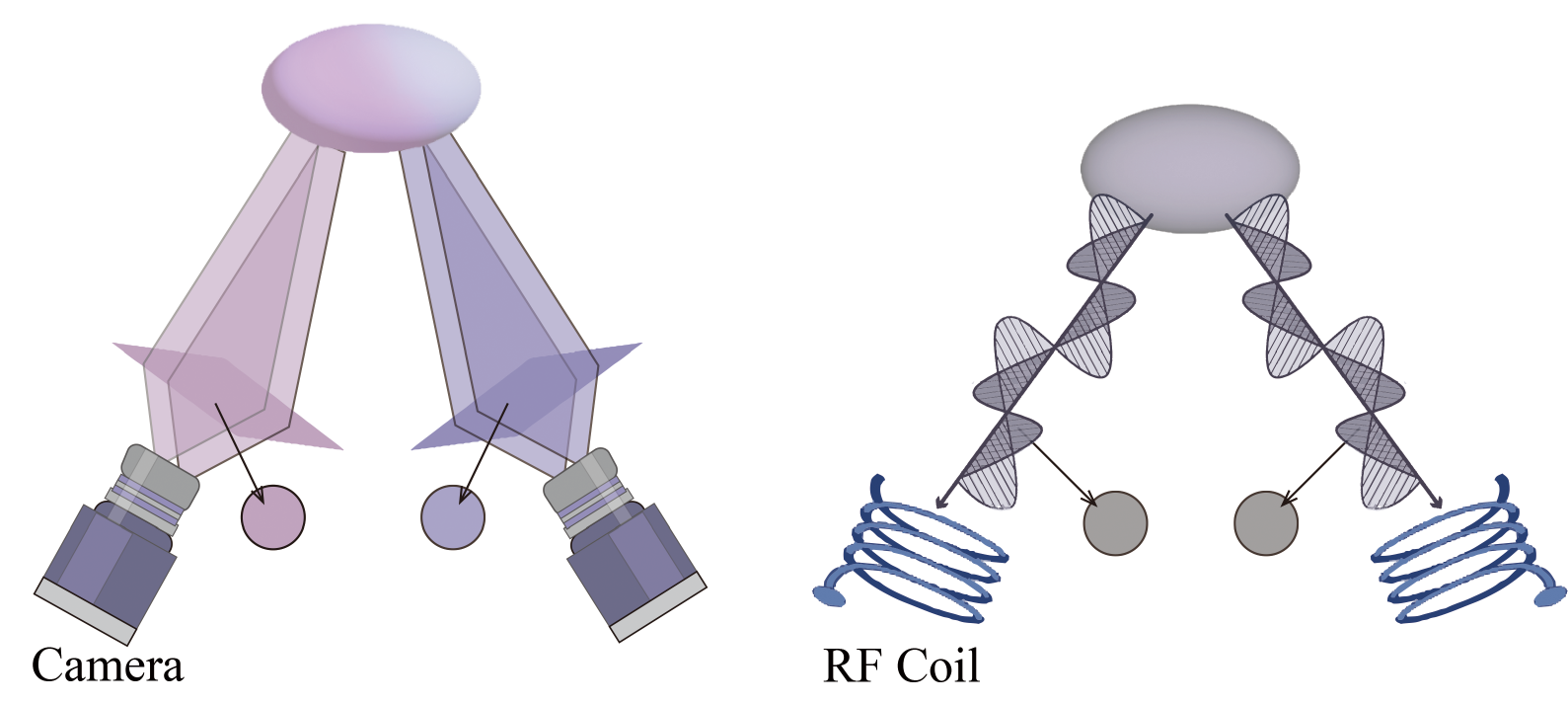}
\caption{Comparison between original 3D Gaussian Splatting primitives and the proposed Magnetic Gaussian primitives. The original 3DGS utilizes view-dependent spherical harmonic coefficients for color representation, whereas M-Gaussian employs a single intensity value to model tissue-specific MRI signal properties, which significantly reduces memory overhead.}
\label{gs_primitive}
\end{figure}

MRI signal acquisition fundamentally differs from optical imaging in both physics and data characteristics. As shown in Fig.~\ref{gs_primitive}, while optical imaging captures surface reflectance that varies with viewing direction, MRI measures intrinsic tissue properties that remain constant regardless of the imaging plane orientation. 

The original 3DGS framework uses spherical harmonics (SH)~\cite{sphericalharmonics} to model view-dependent RGB color $c$,
\begin{equation}
c(\mathbf{v}) = \sum_{l=0}^{L} \sum_{m=-l}^{l} k_{lm} Y_{lm}(\mathbf{v})
\end{equation}
where $\mathbf{v}$ is the view direction, $L$ is the maximum degree of SH coefficients, $Y_{lm}$ are the SH basis functions and $k_{lm}$ are the learnable SH coefficients, requiring $3(L+1)^2$ coefficients per primitive. 
While this is essential for optical rendering, it introduces substantial memory overhead and computational complexity when applied to MRI reconstruction where RGB color and view-dependency are unnecessary. Maintaining view-dependent appearance parameters not only increases memory consumption but also complicates optimization without providing commensurate benefits.

Consequently, we eliminate view-dependent color representation entirely, replacing it with a single scalar intensity value that directly models tissue-specific MRI signal properties.
We propose Magnetic Gaussians, where each primitive $G_i$ is parameterized by:
\begin{equation}
G_i = \{\boldsymbol{\mu}_i, \boldsymbol{\Sigma}_i, \alpha_i\}
\end{equation}
where $\boldsymbol{\mu}_i \in \mathbb{R}^3$ represents the spatial center in normalized coordinates, $\boldsymbol{\Sigma}_i \in \mathbb{R}^{3 \times 3}$ is the covariance matrix encoding the shape and orientation, and $\alpha_i \in [0,1]$ is the normalized MRI signal intensity. By replacing view-dependent spherical harmonics with a single intensity value, our parameterization reduces per-primitive parameters from 59 to 11, achieving 5.4 times memory reduction, which is critical for high-resolution brain imaging where millions of primitives are required.

To ensure numerical stability and efficient gradient-based optimization, we employ a factored representation:
\begin{equation}
\boldsymbol{\Sigma}_i = \mathbf{R}_i \mathbf{S}_i \mathbf{S}_i^T \mathbf{R}_i^T
\end{equation}
where $\mathbf{R}_i \in SO(3)$ is the rotation matrix and $\mathbf{S}_i = \text{diag}(\exp(\mathbf{s}_i)) \in \mathbb{R}^{3 \times 3}$ is the diagonal scale matrix. The rotation matrix $\mathbf{R}_i$ is parameterized using unit quaternions $\mathbf{q}_i \in \mathbb{R}^4$ with $\|\mathbf{q}_i\| = 1$, providing a singularity-free representation for 3D rotations. During training, we optimize unnormalized quaternions and apply normalization before converting to rotation matrices. For the scale matrix, we use learnable log-scale parameters $\mathbf{s}_i = [s_i^x, s_i^y, s_i^z]^T \in \mathbb{R}^3$ with exponential mapping $\mathbf{S}_i = \text{diag}(\exp(\mathbf{s}_i))$. This ensures positive scale values while providing numerically stable gradients.

\subsection{MRI Volume Rendering with Spatial Query}

We designed an efficient rendering pipeline for the proposed M-Gaussian representation. The volumetric nature of MRI reconstruction demands volumetric sampling at arbitrary 3D locations throughout the imaging volume. Direct evaluation of all $N_G$ Gaussians for each query point would result in $O(N_G \times N_{\text{voxels}})$ complexity, becoming computationally prohibitive for high-resolution reconstruction where both terms can reach millions. Our approach is a localized query mechanism whose design is motivated by the physical locality of MRI signals and justified by the mathematically compact support of Gaussian primitives.

\subsubsection{Block-based Spatial Partitioning for Efficient Query}
We introduce a spatial partitioning scheme to accelerate query operations. The normalized volume $[-1, 1]^3$ is partitioned into a uniform grid based on a given $\mathrm{grid\_resolution}$. Each Gaussian $G_i$ is assigned to a grid cell according to its center $\boldsymbol{\mu}_i$:
\begin{equation}
\text{cell\_index}(\boldsymbol{\mu}_i) = \left\lfloor (\boldsymbol{\mu}_i + \mathbf{1}) \cdot \frac{\mathrm{grid\_resolution}}{2} \right\rfloor
\end{equation}

For any query point $\mathbf{x}$, its corresponding cell index, $(i_0, j_0, k_0)$, is first determined. A local neighborhood of cells, $\mathcal{N}_{\text{cells}}(\mathbf{x})$, is then defined within a search radius of $\text{block\_radius}$:
\begin{equation}
\mathcal{N}_{\text{cells}}(\mathbf{x}) = \{\text{cell}_{i,j,k} \mid |i - i_0|, |j - j_0|, |k - k_0| \leq \text{block\_radius}\}
\end{equation}

The active set of Gaussians for rendering, $\mathcal{G}_{\text{local}}$, is subsequently defined as all primitives residing within this cell neighborhood:
\begin{equation}
\mathcal{G}_{\text{local}}(\mathbf{x}) = \{G_i \mid \text{cell\_index}(\boldsymbol{\mu}_i) \in \mathcal{N}_{\text{cells}}(\mathbf{x}) \}
\end{equation}

This partitioning strategy provides fundamental computational advantages over projection-based methods like original 3DGS. While 3DGS requires complex splatting operations involving projection, sorting, and tile-based rasterization, our approach achieves $O(1)$ search time for finding relevant Gaussians through grid-based indexing. Our method eliminates view-dependent computations entirely—traditional 3DGS must re-project and re-sort Gaussians for each viewing angle, whereas our spatial partitioning remains constant regardless of query pattern. This is particularly advantageous for volumetric reconstruction tasks requiring dense 3D space sampling rather than specific 2D view rendering.

\subsubsection{MRI Volumetric Rendering}
Unlike optical imaging where radiance is accumulated along viewing rays, MRI signal generation follows fundamentally different physics. Each voxel in an MRI volume represents the aggregate magnetic resonance signal from tissue within that spatial location. This distinction necessitates a specialized rendering approach that respects the volumetric nature of MRI data while maintaining computational efficiency.

For each sampled point $\mathbf{x}_{\text{sample}}$ from slice $k$, we apply the corresponding rigid transformation to obtain $\mathbf{x} = T_k(\mathbf{x}_{\text{sample}})$ in the reconstructed volume space. These transformations account for inter-slice motion and misalignment inherent in multi-stack acquisitions. Signal intensity is computed by aggregating local Gaussian contributions:
\begin{equation}
I(\mathbf{x}) = \sum_{i \in \mathcal{G}_{\text{local}}(\mathbf{x})} \alpha_i \cdot \exp\left(-\frac{1}{2}(\mathbf{x} - \boldsymbol{\mu}_i)^T \boldsymbol{\Sigma}_i^{-1} (\mathbf{x} - \boldsymbol{\mu}_i)\right)
\end{equation}
The rigid transformations $\{T_k\}$ are jointly optimized with Gaussian parameters during training to achieve accurate slice alignment and volume reconstruction. 
This formulation differs fundamentally from alpha-compositing in NeRF or projection-based rendering in 3DGS. Instead of accumulating along rays or projecting onto planes, we directly evaluate the 3D Gaussian mixture at each query point, aligning with MRI's physical signal formation process. The inherent smoothness of Gaussian basis functions naturally models the Point Spread Function characteristic of MRI acquisition, capturing gradual signal transitions without explicit PSF modeling. Our approach also naturally handles partial volume effects critical in thick-slice reconstruction—when voxels contain multiple tissue types, overlapping Gaussians with different intensities $\alpha_i$ provide principled mixed signal representation, particularly important at tissue boundaries where traditional methods struggle with smooth transitions.

\subsection{Neural Residual Field Enhancement}
Anatomical boundaries between tissues and fine-scale structures exhibit sharp intensity transitions that challenge the inherently smooth Gaussian representation. To address this limitation, we augment the base Gaussian representation with a Neural Residual Field (NRF)—a lightweight MLP that captures high-frequency details beyond the capacity of smooth basis functions.

The NRF employs Fourier positional encoding with 6 frequency bands to map input coordinates $\mathbf{x}$ to higher-dimensional features, enabling the network to learn fine spatial patterns. The architecture consists of 4 hidden layers with 64 neurons each, using SiLU activation functions. The output is bounded through scaled tanh to $[-0.1, 0.1]$:
\begin{equation}
I_{\text{final}}(\mathbf{x}) = I(\mathbf{x}) + r(\mathbf{x})
\end{equation}
where $r(\mathbf{x})$ is the NRF-predicted residual. This bounding ensures residual corrections remain small relative to base Gaussian intensities, preventing overfitting to noise.

NRF is activated after initial Gaussian convergence through a delayed activation strategy. This allows Gaussians to first capture coarse volumetric structures before NRF refines high-frequency details. The delayed activation is critical to prevent the flexible MLP from fitting noise before a robust base representation is established.

\begin{table*}[t]
\centering
\caption{Quantitative comparison with baseline slice-to-volume MRI reconstruction methods across three datasets. Metrics include PSNR, SSIM, NCC, and NRMSE for reconstruction quality, and runtime for computational efficiency. Best results are in \textbf{bold}, second-best are \underline{underlined}.}
\label{tab:comparison}
\begin{tabular}{lccccc}
\hline
Methods & PSNR (dB) $\uparrow$ & SSIM $\uparrow$ & NCC $\uparrow$ & NRMSE $\downarrow$ & Runtime (min) $\downarrow$ \\
\hline
\multicolumn{6}{c}{\textit{FeTA Dataset}} \\
\hline
NiftyMIC~\cite{niftymic}  & \underline{36.08} & 0.9857 & \underline{0.9932} & \underline{0.0157} & 79.12 \\
SVRTK~\cite{svrtk} & 34.64 & \underline{0.9861} & 0.9916 & 0.0185 & \underline{14.75} \\
NeSVoR~\cite{xu2023nesvor} & 31.35 & 0.9588 & 0.9546 & 0.0414 & 31.75 \\
M-Gaussian (Ours) & \textbf{40.31} & \textbf{0.9936} & \textbf{0.9975} & \textbf{0.0096} & \textbf{5.63} \\
\hline
\multicolumn{6}{c}{\textit{FaBiAN Dataset}} \\
\hline
NiftyMIC~\cite{niftymic}  & \underline{32.09} & \textbf{0.9673} & \underline{0.9666} & \underline{0.0249} & 66.15 \\
SVRTK~\cite{svrtk}  & 31.38 & 0.9332 & 0.9497 & 0.0270 & \underline{12.38} \\
NeSVoR~\cite{xu2023nesvor} & 30.15 & 0.9445 & 0.9622 & 0.0311 & 30.42 \\
M-Gaussian (Ours) & \textbf{32.26} & \underline{0.9521} & \textbf{0.9668} & \textbf{0.0244} & \textbf{4.78} \\
\hline
\multicolumn{6}{c}{\textit{HCP Dataset}} \\
\hline
NiftyMIC~\cite{niftymic}  & \textbf{33.34} & \textbf{0.9782} & \underline{0.9917} & \textbf{0.0215} & 1353.48 \\
SVRTK~\cite{svrtk}  & 32.33 & 0.9699 & 0.9869 & 0.0242 & \underline{42.52} \\
NeSVoR~\cite{xu2023nesvor} & 32.79 & 0.9751 & 0.9876 & 0.0229 & 64.88 \\
M-Gaussian (Ours) & \underline{33.10} & \underline{0.9776} & \textbf{0.9970} & \underline{0.0221} & \textbf{17.28} \\
\hline
\end{tabular}
\end{table*}

\subsection{Training}

\subsubsection{Multi-Resolution Progressive Training}
We employ a coarse-to-fine training strategy where the Gaussian grid resolution increases progressively at predefined iteration milestones. During resolution transitions, Gaussian parameters are interpolated from the previous grid using trilinear interpolation for intensity and scale, and normalized linear interpolation for rotation quaternions. This progressive densification provides strong initialization from coarse features and accelerates convergence toward fine-grained anatomical details.

\subsubsection{Loss Function}
Our training objective comprises reconstruction and regularization terms:
\begin{equation}
\mathcal{L} = \mathcal{L}_{\text{recon}} + \mathcal{L}_{\text{reg}}
\end{equation}

The reconstruction loss ensures data fidelity through smooth L1 loss for robust point-wise accuracy and SSIM loss for perceptual quality preservation:
\begin{equation}
\mathcal{L}_{\text{recon}} = \mathcal{L}_{\text{L1}} + \lambda_{\text{SSIM}} \mathcal{L}_{\text{SSIM}}
\end{equation}

We adopt the smooth L1 loss, also known as the Huber loss, which combines the stability of L2 loss near zero with the robustness of L1 loss for larger errors:
\begin{equation}
\text{smooth}_{L1}(x) = \begin{cases}
0.5x^2 & \text{if } |x| < 1 \\
|x| - 0.5 & \text{otherwise}
\end{cases}
\end{equation}
where $x = I_{\text{pred}}(\mathbf{x}) - I_{\text{gt}}(\mathbf{x})$ denotes the difference between predicted and ground truth intensities. This formulation provides quadratic smoothing for small residuals while maintaining linear growth for outliers, making optimization less sensitive to noise.

The regularization loss constrains model behavior:
\begin{equation}
\mathcal{L}_{\text{reg}} = \lambda_{\text{aniso}} \mathcal{L}_{\text{aniso}}
\end{equation}
Anisotropic regularization prevents excessive Gaussian elongation:
\begin{equation}
\mathcal{L}_{\text{aniso}} = \frac{1}{N_G}\sum_i \max\left(0, \frac{\max(\mathbf{s}_i)}{\min(\mathbf{s}_i)} - \lambda_r\right)
\end{equation}
where $\lambda_r$ controls the permissible degree of anisotropy, balancing expressive flexibility with numerical stability.

\begin{figure*}[t]
\centering
\includegraphics[width=0.8\textwidth]{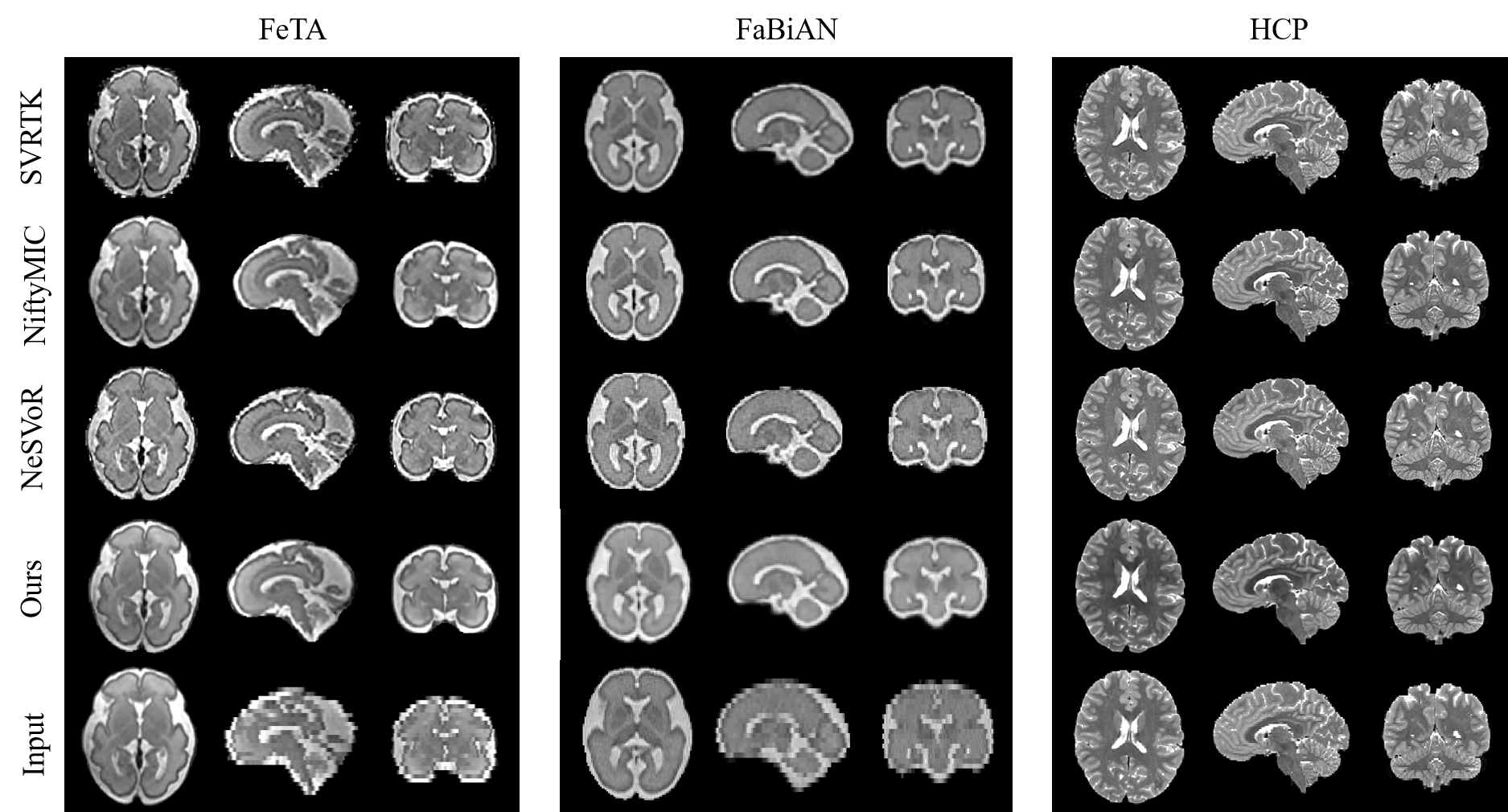}
\caption{Qualitative comparison of slice-to-volume MRI reconstruction methods. Each row shows results from a different dataset (FeTA, FaBiAN, HCP). The Input column displays one representative stack from the multi-stack acquisition. Our M-Gaussian method produces reconstructions with sharper anatomical boundaries and fewer artifacts compared to baseline methods.}
\label{fig:qualitative}
\end{figure*}

\section{Experiments}
\label{sec:experiments}

\subsection{Experimental Settings}

\subsubsection{Datasets}
We evaluate our method on three datasets:

\begin{itemize}
\item FeTA~\cite{feta}: Fetal brain MRI volumes with 0.5 mm isotropic resolution. We simulate clinical thick-slice acquisitions by downsampling to $0.8\times0.8\times3$ mm$^3$ with added motion artifacts.

\item FaBiAN~\cite{zendo}: Synthetic fetal brain dataset generated at $1.1\times1.1\times3$ mm$^3$ resolution with k-space noise and stochastic inter-slice motion. Controlled ground truth is provided for quantitative evaluation.

\item HCP~\cite{hcp}: Adult brain MRI from the Human Connectome Project at 0.7 mm isotropic resolution, downsampled to $1.0\times1.0\times2$ mm$^3$ to simulate clinical acquisitions. 
\end{itemize}

For all datasets, three orthogonal stacks (axial, coronal, sagittal) are generated per volume to simulate the clinical multi-stack acquisition protocol. The selection of these diverse datasets allows for a comprehensive evaluation across different anatomical scales, age groups, and acquisition characteristics, ensuring the robustness of the proposed method in various clinical scenarios.

\subsubsection{Baselines}
We compare against three representative methods: SVRTK~\cite{svrtk}, a toolkit employing iterative reconstruction with robust statistics; NiftyMIC~\cite{niftymic}, a comprehensive pipeline integrating motion correction, bias field estimation, and intensity standardization; and NeSVoR~\cite{xu2023nesvor}, an implicit neural representation method. For fair comparison, NeSVoR is evaluated using its pure PyTorch~\cite{pytorch} implementation without custom CUDA kernels.

\subsubsection{Evaluation Metrics}
Reconstruction quality is assessed using Peak Signal-to-Noise Ratio (PSNR), Structural Similarity Index (SSIM), Normalized Cross-Correlation (NCC), and Normalized Root Mean Square Error (NRMSE). Runtime (in minutes) is also reported to evaluate computational efficiency.

\subsubsection{Implementation Details}
M-Gaussian is implemented in PyTorch. Gaussian parameters are optimized using Adam with learning rates of 0.001, 0.05, 0.005, and 0.001 for position, intensity, scale, and rotation, respectively. The Neural Residual Field employs 6 Fourier frequency bands with 4 hidden layers of 64 neurons each, optimized with Adam at a learning rate of 0.0001. The NRF is activated at iteration 2000 for FeTA/FaBiAN and iteration 4000 for HCP.
Loss weights are set to $\lambda_{\text{SSIM}}=0.5$, $\lambda_{\text{aniso}}=0.1$, and $\lambda_r=1.5$. The default block radius for spatial querying is set to 5.

Target resolutions are 0.5 mm (FeTA), 1.1 mm (FaBiAN), and 0.7 mm (HCP). Progressive training starts from $70^3$ grids, increasing to $200^3$ for FeTA/FaBiAN at iterations 500, 1000, 2000, and 3000. HCP follows an extended schedule reaching $280^3$ at iteration 6000 to capture finer anatomical structures. Total training iterations are 4000 for FeTA/FaBiAN and 8000 for HCP. All experiments are conducted on a system with 2 AMD EPYC 7352 CPUs and an NVIDIA RTX A6000 GPU (48 GB memory).

\subsection{Comparison with Baseline Methods}

Table~\ref{tab:comparison} presents quantitative comparisons across the three datasets, and Fig.~\ref{fig:qualitative} illustrates representative qualitative results. Our method achieves an optimal balance between reconstruction quality and computational efficiency across all evaluated scenarios.

On the \textit{FeTA} dataset, M-Gaussian achieves a PSNR of 40.31 dB, outperforming the second-best method (NiftyMIC) by a substantial margin of 4.23 dB. Notably, this quality improvement is accompanied by a 14$\times$ speedup in runtime (5.63 min vs. 79.12 min). The superior performance on fetal data demonstrates our method's effectiveness in handling the challenging scenario of small brain volumes with limited anatomical context.

The efficiency advantage becomes more pronounced on the high-resolution \textit{HCP} dataset. M-Gaussian completes reconstruction in just 17.28 minutes—a 78$\times$ acceleration compared to NiftyMIC (1353.48 min)—while maintaining competitive accuracy with the highest NCC and second-best PSNR. This demonstrates the scalability of our approach to high-resolution adult brain reconstruction.

On the \textit{FaBiAN} dataset, M-Gaussian achieves the highest PSNR and NCC in under 5 minutes. While NiftyMIC achieves slightly higher SSIM, our method provides a better trade-off considering the nearly 14$\times$ faster runtime.

\begin{figure}[b]
\centering
\includegraphics[width=0.9\columnwidth]{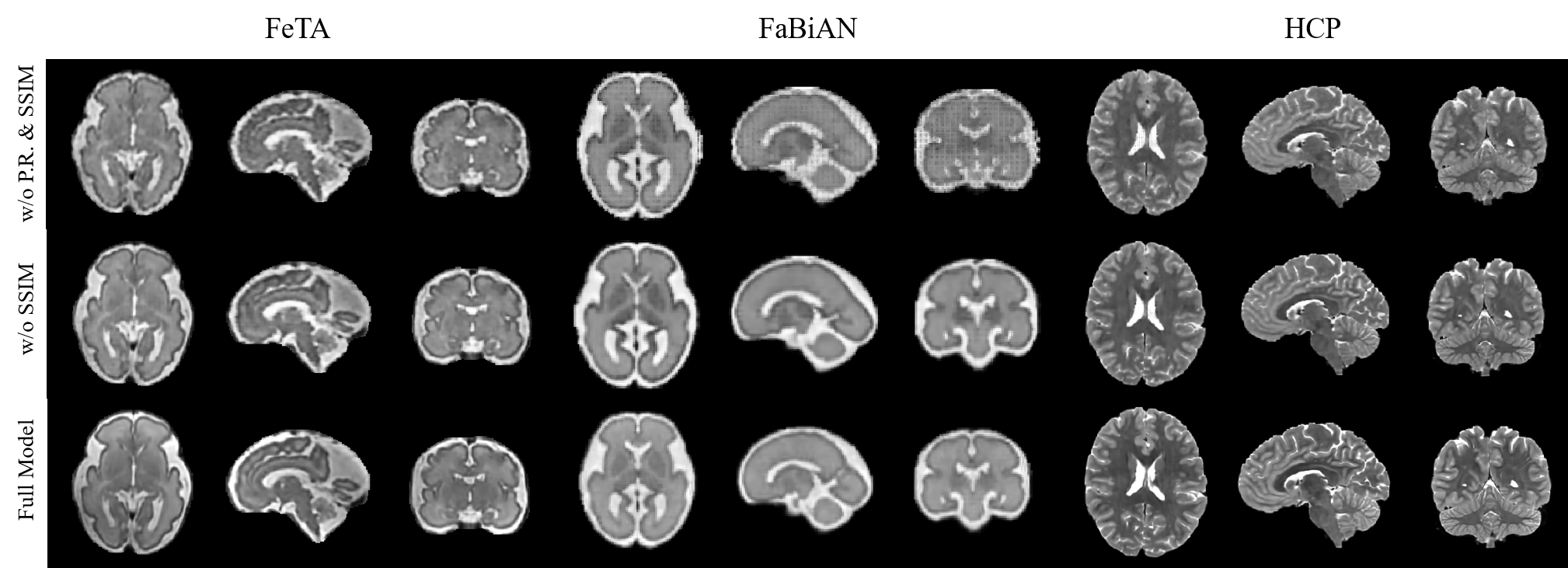}
\caption{Qualitative ablation results on Progressive resolution (P.R.) training and SSIM loss. Progressive resolution training stabilizes convergence, while the inclusion of SSIM loss significantly enhances the preservation of structural integrity and contrast. The effect is particularly significant on FeTA dataset.}
\label{fig:ablation_ssimpr}
\end{figure}

Qualitatively, as shown in Fig.~\ref{fig:qualitative}, M-Gaussian produces reconstructions with sharper tissue boundaries and more coherent anatomical structures. The baseline methods exhibit varying degrees of blurring (NiftyMIC, SVRTK) or noise (NeSVoR), particularly in regions with complex anatomy or near tissue interfaces.

\subsection{Ablation Study}

We conduct systematic ablation experiments to evaluate the contribution of each proposed component. Table~\ref{tab:ablation} reports quantitative results.

\subsubsection{Progressive Resolution Training}
We compare the proposed multi-resolution progressive training schedule against a baseline trained at fixed high resolution. To isolate the effect of progressive resolution, both configurations are optimized without SSIM loss, as SSIM evaluation requires sampling entire slices and would introduce confounding factors in analyzing convergence behavior.

As shown in Table~\ref{tab:ablation}, progressive training improves PSNR by 1.17 dB, 2.58 dB, and 1.20 dB on FeTA, FaBiAN, and HCP respectively. Fig.~\ref{fig:ablation_ssimpr} further demonstrates that progressive resolution training substantially accelerates convergence and yields higher-quality reconstructions. Direct optimization of millions of primitives at full resolution leads to unstable dynamics, whereas the coarse-to-fine strategy stabilizes training by gradually increasing representational complexity.

\subsubsection{Structural Similarity Loss}
To assess the role of the SSIM loss, we contrast the full model with a variant trained using only L1 reconstruction loss. As shown in Table~\ref{tab:ablation}, incorporating SSIM loss yields dramatic improvements: 8.07 dB on FeTA, 1.25 dB on FaBiAN, and 0.54 dB on HCP. Fig.~\ref{fig:ablation_ssimpr} demonstrates that SSIM loss enhances preservation of anatomical boundaries and tissue contrast, maintaining clinically relevant morphology that pixel-wise losses alone cannot capture.

\begin{figure}[t]
\centering
\includegraphics[width=0.9\columnwidth]{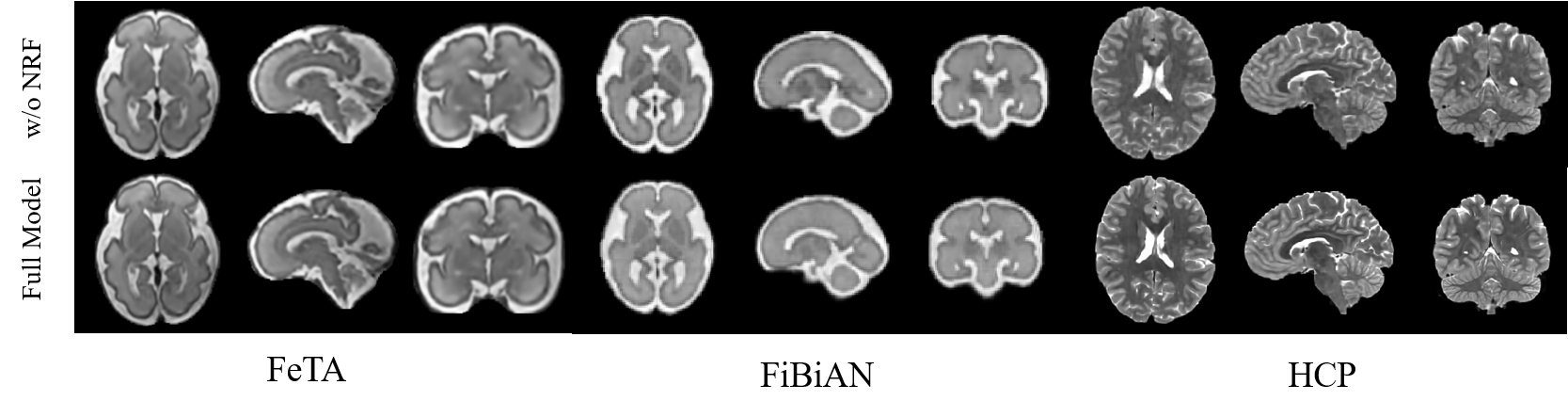}
\caption{Qualitative ablation results on Neural Residual Field (NRF). The NRF suppresses noise and refines high-frequency details, which is prominent on the HCP dataset.}
\label{fig:ablation_nrf}
\end{figure}

\subsubsection{Neural Residual Field Enhancement}
We evaluate the NRF contribution by comparing the full model against a variant without neural residual refinement. Removing NRF results in PSNR drops of 1.20 dB (FeTA), 1.33 dB (FaBiAN), and 1.14 dB (HCP). Fig.~\ref{fig:ablation_nrf} shows that the variant without NRF exhibits increased noise, particularly near tissue boundaries. We attribute this to the inherent smoothness of Gaussian basis functions, which struggle to model sharp intensity transitions. The NRF addresses this limitation by providing a complementary representation well-suited for high-frequency patterns, enabling Gaussians to focus on coarse-scale modeling while NRF refines fine details.

\begin{figure}[h]
\centering
\includegraphics[width=0.9\columnwidth]{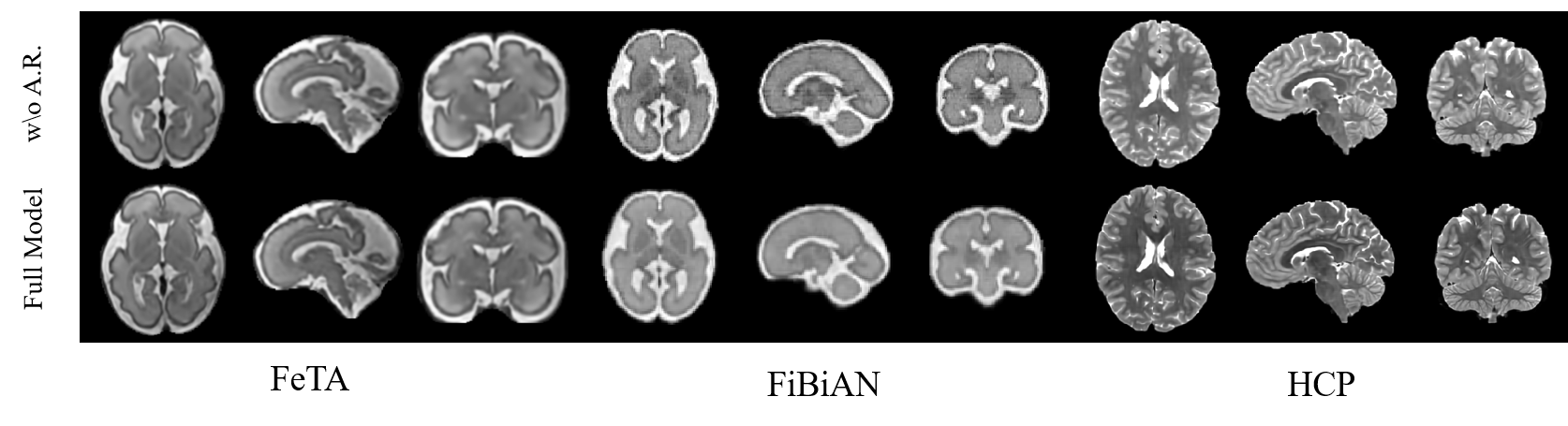}
\caption{Qualitative ablation results on Anisotropic Regularization (A.R.). The absence of A.R. leads to needle-like artifacts due to degenerate Gaussian geometries. These artifacts are effectively mitigated in the full model, which is most pronounced on the FaBiAN dataset.}
\label{fig:ablation_ar}
\end{figure}

\subsubsection{Anisotropic Regularization}
Disabling anisotropic regularization during training results in PSNR degradation of 0.82 dB (FeTA), 2.24 dB (FaBiAN), and 0.28 dB (HCP). As illustrated in Fig.~\ref{fig:ablation_ar}, the absence of this regularizer leads to streaking and granular noise artifacts. We attribute this to the emergence of needle-like Gaussians with extreme aspect ratios during optimization. The regularization prevents such degenerate geometries by penalizing aspect ratios exceeding $\lambda_r$, ensuring Gaussians maintain moderate anisotropy conducive to smooth volumetric reconstruction.

\begin{table}[t]
\centering
\caption{Ablation study quantifying the contribution of structural similarity loss (SSIM), progressive resolution training (P.R.), Neural Residual Field (NRF), and anisotropic regularization (A.R.).}
\label{tab:ablation}
\scriptsize
\begin{tabular}{l|cc|cc|cc}
\hline
\multirow{2}{*}{Configuration} & \multicolumn{2}{c|}{\textbf{FeTA}} & \multicolumn{2}{c|}{\textbf{FaBiAN}} & \multicolumn{2}{c}{\textbf{HCP}} \\
& PSNR & SSIM & PSNR & SSIM & PSNR & SSIM \\
\hline
Full model & 40.31 & 0.9936 & 32.26 & 0.9521 & 33.10 & 0.9776 \\
w/o SSIM & 32.24 & 0.9611 & 31.01 & 0.9388 & 32.56 & 0.9582 \\
w/o SSIM \& P.R. & 31.07 & 0.9573 & 28.43 & 0.9272 & 31.36 & 0.9512 \\
w/o NRF & 39.11 & 0.9893 & 30.93 & 0.9387 & 31.96 & 0.9538 \\
w/o A.R. & 39.49 & 0.9902 & 30.02 & 0.9486 & 32.82 & 0.9606 \\
\hline
\end{tabular}
\end{table}

\subsection{Hyperparameter Analysis}

Beyond validating the necessity of each component, we investigate how key hyperparameters affect the quality-efficiency trade-off. 

\subsubsection{Gaussian Grid Resolution}

The Gaussian grid resolution directly determines the model's representational capacity and computational requirements. To enable direct comparison across different resolutions, we disable 
progressive training and train the model with fixed resolutions ranging from $50^3$ to $400^3$ on the HCP dataset. Resolutions beyond $400^3$ could not be evaluated due to GPU memory constraints.

As shown in Table~\ref{tab:resolution}, reconstruction quality improves substantially as resolution increases from $50^3$ to $150^3$, with PSNR rising from 30.27 dB to 31.31 dB (+1.04 dB). At higher resolutions, however, the gains become marginal—from $150^3$ to $350^3$, PSNR improves by only 0.45 dB while training time increases by 65\%. Notably, performance at $400^3$ (31.63 dB) drops slightly below that of $350^3$ (31.76 dB), indicating that excessively high resolution without progressive training may lead to optimization difficulties or overfitting.

These findings strongly motivate our progressive training strategy. Comparing with results in Table~\ref{tab:comparison}, our full model with progressive training achieves 33.10 dB on HCP in 17.28 minutes—outperforming the best fixed-resolution result (31.76 dB at $350^3$) by 1.34 dB while being 36\% faster. The coarse-to-fine optimization path not only improves efficiency but also enables better final reconstruction quality by providing stable gradient flow and strong initialization at each resolution stage.

\subsubsection{Effect of Block Radius}

The block radius parameter controls the spatial extent of Gaussian contributions considered for each query point, directly impacting the trade-off between computational efficiency and reconstruction accuracy. A smaller radius reduces the number of Gaussians evaluated per query but may exclude relevant contributions, while a larger radius increases computational cost and may cause over-smoothing by aggregating excessive Gaussian responses.

\begin{table}[h]
\centering
\caption{Impact of Gaussian grid resolution on reconstruction fidelity and training time using the HCP dataset. The models were trained with fixed resolutions, excluding the effect of progressive training strategy.}
\label{tab:resolution}
\begin{tabular}{cccc}
\hline
Resolution & PSNR $\uparrow$ & SSIM $\uparrow$ & Time (min) $\downarrow$ \\
\hline
$50^3$ & 30.27 & 0.9352 & 10.82 \\
$150^3$ & 31.31 & 0.9505 & 16.43 \\
$250^3$ & 31.50 & 0.9527 & 24.70 \\
$350^3$ & 31.76 & 0.9547 & 27.07 \\
$400^3$ & 31.63 & 0.9546 & 30.30 \\
\hline
\end{tabular}
\end{table}


\begin{figure}[htbp]
  \centering
  
  \begin{minipage}{0.48\textwidth}
    \centering
    \includegraphics[width=\textwidth]{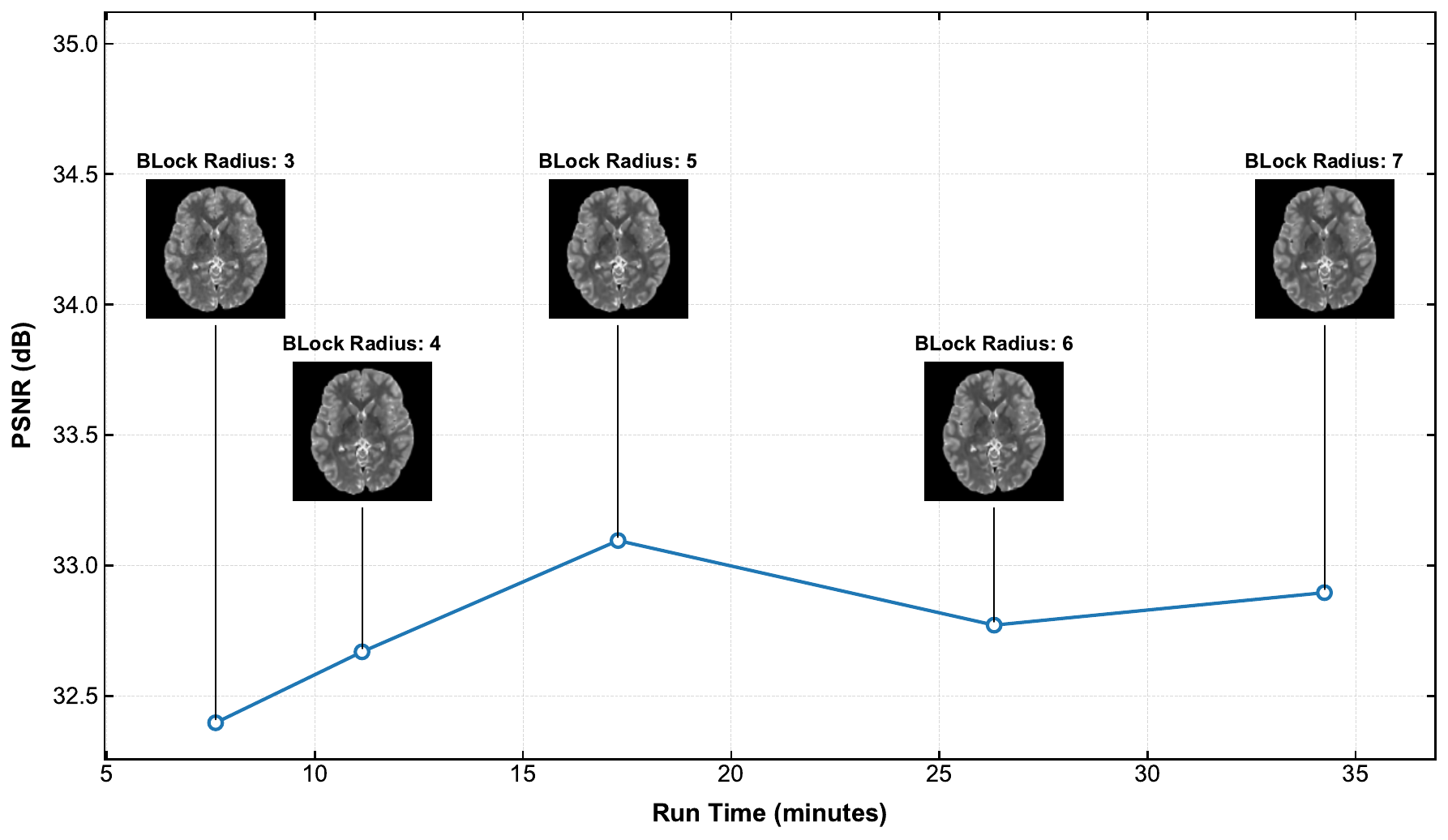}
    \caption{Analysis of block radius on reconstruction quality and training time. A block radius of 5 achieves the optimal balance, maximizing PSNR while maintaining computational efficiency.}
    \label{fig:br_psnr_time}
  \end{minipage}
  \hfill 
  \begin{minipage}{0.48\textwidth}
    \centering
    \includegraphics[width=\textwidth]{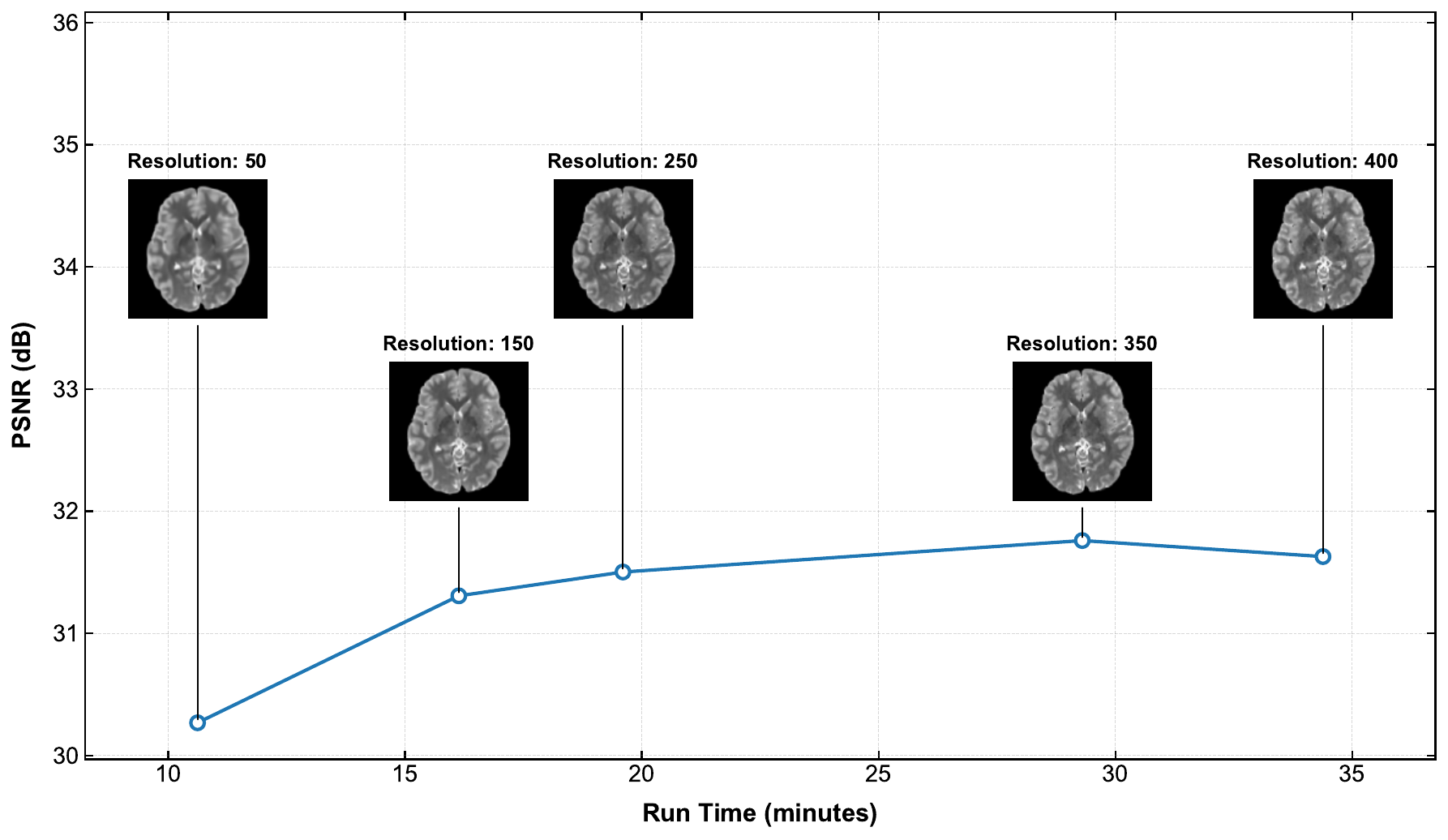}
    \caption{Analysis of Gaussian grid resolution on reconstruction quality and training time. While higher resolutions improve quality, the performance plateaus beyond $350^3$.}
    \label{fig:reso_psnr_time}
  \end{minipage}

\end{figure}

Table~\ref{tab:blockradius} summarizes the results across block radius values from 3 to 7. Block radius 5 achieves the highest PSNR (33.10 dB), surpassing both smaller and larger values. This non-monotonic behavior reflects the dual nature of the radius parameter. When the radius is too small, relevant Gaussian contributions may be excluded, and the limited number of Gaussians in the aggregation makes the rendered intensity more susceptible to fluctuations, introducing noise. Conversely, when the radius is too large, distant Gaussians with negligible contributions are included, and the excessive aggregation tends to over-smooth fine anatomical details.

The efficiency gap across configurations is also substantial: radius 3 completes training in 7.62 minutes while radius 7 requires 34.24 minutes—a 4.5$\times$ difference. Given that radius 5 achieves optimal quality while maintaining moderate training time, we adopt it as the default setting. This choice balances reconstruction accuracy with computational efficiency, making M-Gaussian practical for routine clinical deployment.

\begin{table}[h]
\centering
\caption{Quantitative assessment of reconstruction performance under different block radius settings. The results illustrate the trade-off between reconstruction quality and runtime.}
\label{tab:blockradius}
\begin{tabular}{cccc}
\hline
Block Radius & PSNR  $\uparrow$ & SSIM $\uparrow$ & Time (min) $\downarrow$ \\
\hline
3 & 32.40 & 0.9733& 7.62 \\
4 & 32.67 & 0.9768& 11.13 \\
5 & 33.10 & 0.9776& 17.28 \\
6 & 32.77 & 0.9778& 26.32 \\
7 & 32.89 & 0.9780& 34.24 \\
\hline
\end{tabular}
\end{table}

\begin{table*}[t]
\centering
\caption{Quantitative evaluation of segmentation performance on the FeTA dataset. Metrics include Dice, HD95, and ASSD. Best results are \textbf{bold}, second-best are \underline{underlined}.}
\label{tab:merged_segmentation_vlines}
\scriptsize
\setlength{\tabcolsep}{3.5pt}

\begin{tabular}{l|cccc|cccc|cccc}
\hline
\multirow{2}{*}{Structure} & \multicolumn{4}{c|}{Dice $\uparrow$} & \multicolumn{4}{c|}{HD95 (mm) $\downarrow$} & \multicolumn{4}{c}{ASSD (mm) $\downarrow$} \\
\cline{2-13}

 & NiftyMIC & SVRTK & NeSVoR & M-Gaussian & NiftyMIC & SVRTK & NeSVoR & M-Gaussian & NiftyMIC & SVRTK & NeSVoR & M-Gaussian \\
\hline
External CSF & 0.859 & 0.883 & \underline{0.888} & \textbf{0.905} & 1.43 & 1.42 & \underline{1.10} & \textbf{0.97} & 0.55 & 0.53 & \underline{0.46} & \textbf{0.41} \\
Grey Matter  & 0.832 & 0.816 & \underline{0.843} & \textbf{0.846} & 0.82 & 0.85 & \underline{0.68} & \textbf{0.66} & 0.37 & 0.38 & \underline{0.34} & \textbf{0.32} \\
White Matter & 0.945 & 0.937 & \underline{0.951} & \textbf{0.953} & 0.95 & 1.11 & \underline{0.80} & \textbf{0.69} & 0.39 & 0.46 & \underline{0.36} & \textbf{0.34} \\
Ventricles   & \underline{0.914} & 0.818 & 0.906 & \textbf{0.917} & \underline{0.64} & 1.23 & 0.65 & \textbf{0.63} & \textbf{0.28} & 0.52 & 0.30 & \textbf{0.28} \\
Cerebellum   & 0.924 & \underline{0.934} & \textbf{0.941} & 0.931 & 1.12 & \underline{0.94} & \textbf{0.87} & 1.05 & 0.46 & \underline{0.40} & \textbf{0.36} & 0.42 \\
Deep GM      & 0.899 & 0.830 & \textbf{0.922} & \underline{0.920} & \underline{1.50} & 3.34 & \textbf{1.10} & 1.52 & \underline{0.56} & 1.00 & \textbf{0.45} & 0.57 \\
Brainstem    & 0.902 & 0.776 & \underline{0.922} & \textbf{0.925} & 1.10 & 3.57 & \underline{0.86} & \textbf{0.83} & 0.45 & 0.99 & \underline{0.37} & \textbf{0.35} \\
\hline
\textbf{Mean} & 0.896 & 0.856 & \underline{0.910} & \textbf{0.914} & 1.08 & 1.78 & \textbf{0.87} & \underline{0.91} & 0.44 & 0.61 & \textbf{0.38} & \textbf{0.38} \\
\hline
\end{tabular}
\end{table*}


\subsection{Evaluation of Downstream Segmentation Tasks}
\label{sec:downstream}

To validate the clinical utility of the reconstructed volumes, we evaluated their performance in a downstream automated brain segmentation task. While standard image quality metrics such as PSNR and SSIM quantify signal fidelity, segmentation accuracy provides a more clinically meaningful assessment of the extent to which semantic anatomical boundaries are preserved for subsequent analysis.

\subsubsection{Experimental Protocol}
We employed nnU-Net~\cite{nnunet}, a well-established deep learning framework for medical image segmentation widely adopted as a benchmark in the field. Our evaluation follows a domain generalization protocol designed to rigorously assess reconstruction quality: a 3D nnU-Net model is first trained exclusively on ground-truth isotropic high-resolution volumes from the FeTA dataset, ensuring the segmentation network learns optimal anatomical features from artifact-free reference data. This pre-trained model is then applied directly to volumes reconstructed by each method without any fine-tuning. Since the segmentation network has not been exposed to reconstruction artifacts during training, any performance degradation directly reflects geometric distortions, blurring, or boundary inconsistencies introduced by the reconstruction process. Seven anatomically distinct structures are segmented: external cerebrospinal fluid (CSF), grey matter, white matter, ventricles, cerebellum, deep grey matter (Deep GM), and brainstem.

\subsubsection{Volumetric Overlap Analysis}
Table~\ref{tab:merged_segmentation_vlines} presents Dice score quantifying volumetric overlap between predicted and ground-truth segmentations. M-Gaussian achieves the highest mean Dice, surpassing NeSVoR and substantially outperforming traditional methods NiftyMIC and SVRTK. Performance analysis reveals distinct patterns correlating with anatomical scale. For large, spatially extensive structures such as external CSF, white matter, and grey matter, M-Gaussian demonstrates particularly strong performance, as these regions benefit from the Gaussian primitives' capacity to efficiently model smooth volumetric distributions. In contrast, for smaller, geometrically complex structures including cerebellum and deep grey matter, the performance gap between M-Gaussian and NeSVoR narrows considerably, suggesting that implicit neural representations may retain marginal advantages for highly localized structures requiring dense spatial sampling. Despite these structure-specific variations, M-Gaussian maintains the highest overall mean Dice across all seven anatomical regions.

\subsubsection{Boundary Localization Accuracy}
Beyond volumetric agreement, precise boundary localization is essential for clinical applications requiring geometric measurements, such as cortical thickness analysis. Table~\ref{tab:merged_segmentation_vlines} reports 95th percentile Hausdorff Distance (HD95) and Average Symmetric Surface Distance (ASSD). M-Gaussian achieves the lowest boundary errors for the majority of structures. Notably, for grey matter and white matter—regions characterized by highly convoluted cortical folding—our method demonstrates notable improvements over NeSVoR in HD95. For external CSF, which demarcates the challenging low-contrast interface between brain tissue and surrounding fluid, M-Gaussian also achieves considerably lower HD95 compared to both NeSVoR and traditional methods. The superior boundary fidelity can be attributed to the complementary architecture: while Gaussian primitives efficiently model smooth volumetric intensity distributions, the Neural Residual Field captures high-frequency spatial variations localized at tissue boundaries.

\begin{figure}[t]
\centering
\includegraphics[width=0.8\columnwidth]{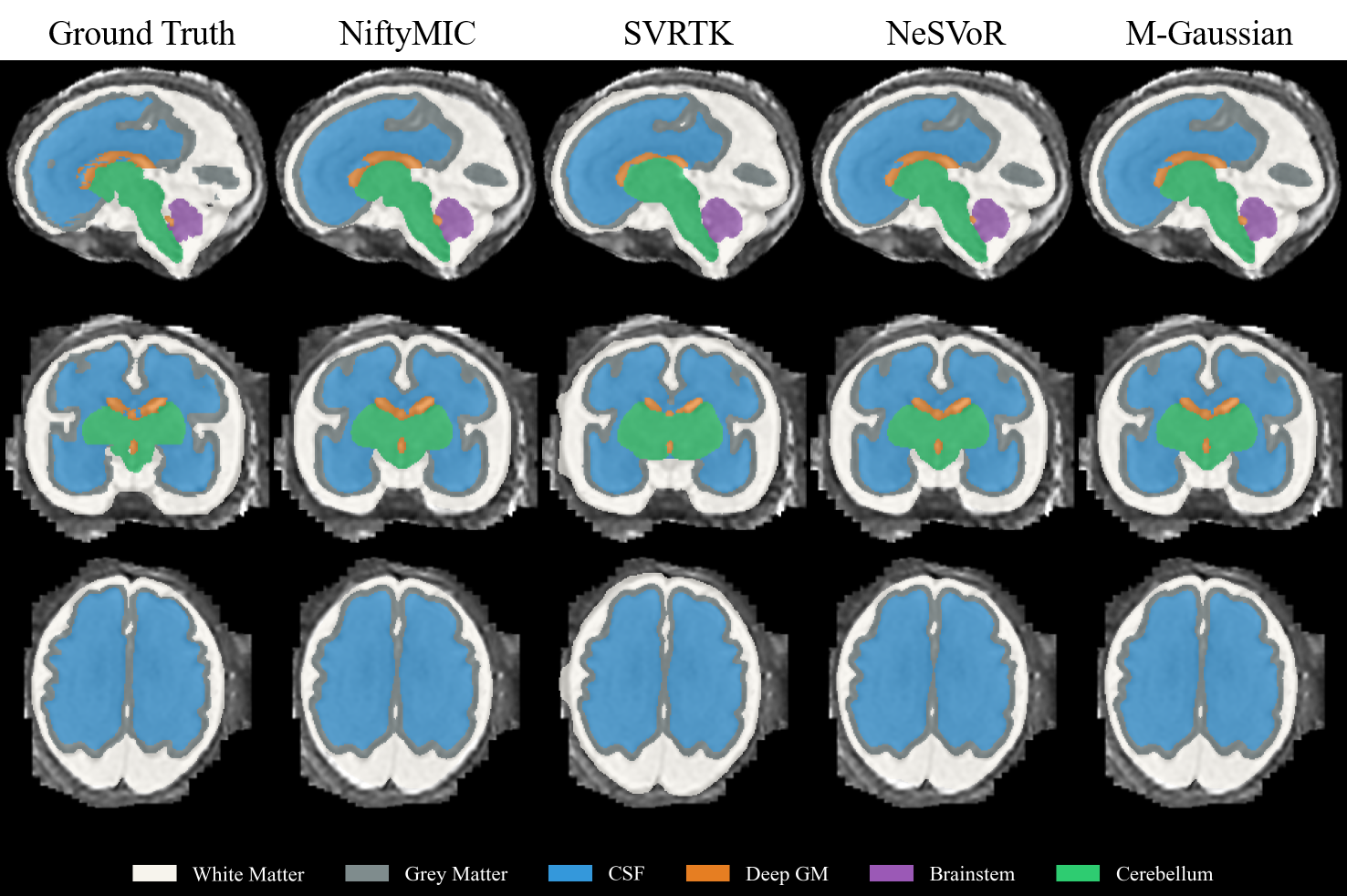}
\caption{Qualitative comparison of downstream segmentation results on the FeTA dataset. M-Gaussian reconstructions yield segmentation masks with the closest correspondence to ground truth, particularly for large-scale structures such as the cortical ribbon and ventricular boundaries.}
\label{fig:segmentation}
\end{figure}

Figure~\ref{fig:segmentation} provides qualitative validation across three orthogonal planes. The segmentation masks derived from M-Gaussian reconstructions exhibit the closest visual correspondence with ground truth, particularly for large-scale structures such as the cortical ribbon and ventricular system. For finer structures including the cerebellum and brainstem, all learning-based methods achieve comparable delineation accuracy, consistent with the quantitative findings.

\section{Discussion}
\label{sec:discussion}

The experimental results demonstrate that M-Gaussian achieves a favorable balance between reconstruction quality and computational efficiency. Several aspects of our approach merit further discussion.

The success of M-Gaussian can be attributed to the synergy between explicit Gaussian representations and neural refinement. Gaussians efficiently capture smooth volumetric structures through direct evaluation, while the Neural Residual Field complements this by modeling high-frequency details that challenge smooth Gaussian basis functions. Block-based spatial partitioning reduces computational complexity from quadratic to near-linear, and progressive training accelerates convergence through strong coarse-scale initialization.

The downstream segmentation evaluation provides additional validation beyond standard image quality metrics. M-Gaussian achieves superior performance on large-scale anatomical structures while showing comparable results to implicit methods on smaller, geometrically complex regions. Large structures with relatively smooth intensity distributions align naturally with the Gaussian basis functions, whereas fine structures requiring dense local sampling may benefit from the continuous querying capability of implicit representations. Nevertheless, the overall segmentation accuracy demonstrates that M-Gaussian reconstructions preserve clinically relevant anatomical boundaries, supporting its potential utility in automated analysis pipelines.

From a clinical workflow perspective, the substantial acceleration achieved by M-Gaussian addresses a critical bottleneck in current reconstruction pipelines. Traditional optimization-based methods and recent implicit neural approaches often require processing times measured in minutes to hours, limiting their applicability in time-sensitive clinical scenarios. The reduced computational burden enables more practical integration into clinical workflows, potentially facilitating real-time quality assessment during acquisition and rapid turnaround for diagnostic interpretation.

It is noteworthy that the current framework assumes rigid inter-slice motion, which may not hold in cases of severe fetal movement or organ deformation. Future directions include extending the motion model to handle non-rigid deformations and exploring joint optimization of acquisition and reconstruction protocols.

\section{Conclusion} 
\label{sec:conclusion}
In this work, we introduced M-Gaussian, the first successful adaptation of 3D Gaussian Splatting for MRI slice-to-volume reconstruction. By reformulating Gaussian primitives to model intrinsic tissue properties and developing a volumetric rendering pipeline consistent with MRI physics, M-Gaussian successfully transformed 3DGS, originally designed for optical imaging, into the field of MRI imaging. The proposed block-based spatial partitioning enables efficient volumetric queries, while the Neural Residual Field captures fine anatomical details beyond the capacity of smooth Gaussian representations. Our multi-resolution progressive training strategy ensures stable convergence for high-resolution reconstruction.

Experimental results on three diverse datasets demonstrate that M-Gaussian achieves an optimal balance between reconstruction quality and runtime efficiency. On the FeTA dataset, our method achieves 40.31 dB PSNR while being 14$\times$ faster than competing methods. On the high-resolution HCP dataset, M-Gaussian provides 78$\times$ acceleration compared to NiftyMIC while maintaining competitive accuracy. These results establish the potential of explicit Gaussian-based representations as an efficient alternative for MRI image reconstruction, opening new avenues for real-time clinical applications.

\bibliographystyle{IEEEtran}
\bibliography{tmi2026}

\end{document}